# A Fourier-enhanced multi-modal 3D small object optical mark recognition and positioning method for percutaneous abdominal puncture surgical navigation

Zezhao Guo[1], Yanzhong Guo[+], Zhanfang Zhao[*]

[1]College of information and Engineering, Hebei GEO University
[2]Beijing Yingrui Pioneer Medical Technology Co., Ltd

## Abstract

Navigation for thoracoabdominal puncture surgery is used to locate the needle entry point on the patient's body surface. The traditional reflective ball navigation method is difficult to position the needle entry point on the soft, irregular, smooth chest and abdomen. Due to the lack of clear characteristic points on the body surface using structured light technology, it is difficult to identify and locate arbitrary needle insertion points. Based on the high stability and high accuracy requirements of surgical navigation, this paper proposed a novel method, a muti-modal 3D small object medical marker detection method, which identifies the center of a small single ring as the needle insertion point. Moreover, this novel method leverages Fourier transform enhancement technology to augment the dataset, enrich image details, and enhance the network's capability. The method extracts the Region of Interest (ROI) of the feature image from both enhanced and original images, followed by generating a mask map. Subsequently, the point cloud of the ROI from the depth map is obtained through the registration of ROI point cloud contour fitting. In addition, this method employs Tukey loss for optimal precision. The experimental results show this novel method proposed in this paper not only achieves high-precision and high-stability positioning, but also enables the positioning of any needle insertion point.

## 1. Introduction

Navigation for thoracoabdominal puncture surgery is used to locate the needle entry point on the patient's body surface. The traditional reflective ball navigation method is difficult to position the needle entry point on the soft, irregular and smooth chest and abdomen. Due to the lack of clear feature points on the body surface using structured light technology, it is difficult to identify and locate any needle insertion point. In order to solve this problem, we use a white 24mm diameter hollow ring to identify medical markers. The ceramic material can be imaged well under CT, and has excellent color-based optical recognition and tomography-based imaging recognition performance. It can fit well on the human body surface and accurately position itself in the body surface environment. However, 3D small object detection is an industry challenge.

* *Corresponding author.   +equal contribution.
E-mail address: zhaozhanfang@hgu.edu.cn

3D object detection has been widely used in actual scenes. With the progress of deep learning technology in point clouds[1][2][3][4], 3D object detection methods have shown significant progress[5][6][7][8]. However, small object detection is still a challenge in the 3D field. In autonomous driving scenarios[9], the performance gap between cars and pedestrians is obvious. In indoor scenes[10][11], the difference in size is larger (e.g. a wardrobe is 1000 times bigger than a cup), so the situation is even worse. For indoor 3D object detection, although the speed and accuracy of furniture-level benchmarks[10][12][13] have been greatly improved, they are still far from practical applications due to the limited range of object sizes they can handle. For example, it is difficult for previous methods to detect small desktop objects[14].

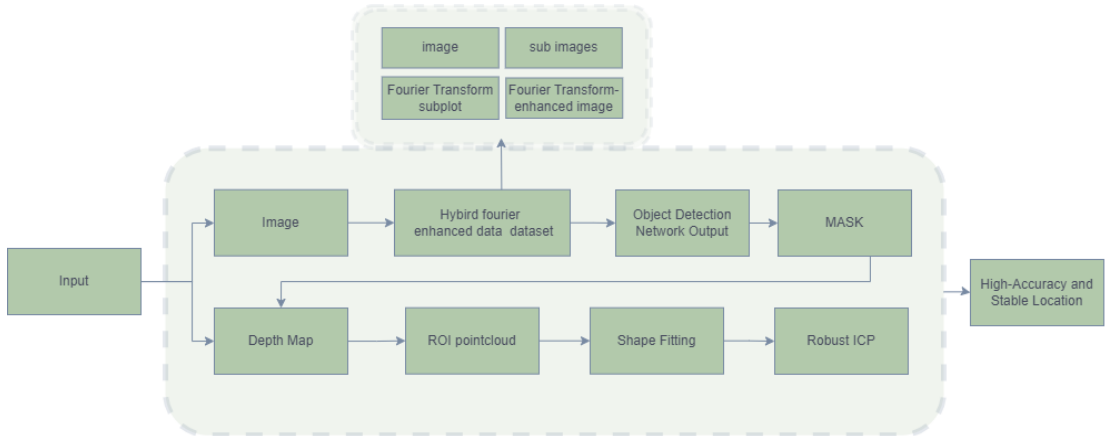

*Figure 1: Image processing and object detection flow chart. The input data includes images and depth maps, which are successively processed by Fourier transform enhancement and ROI masking of the object detection network and point cloud refinement in the second step. First, Fourier transform is used for dataset enhancement. The enhanced dataset is fed into the object detection network, and the detection results are generated to perform mask operations to extract the object and generate ROI areas. The depth map is then used for shape fitting and a robust iterative closest point (ICP) algorithm to Achieve high-precision and stable object positioning.*

This paper proposes a multi-modal fusion perception method of Fourier enhancement, as shown in Figure 1. Fast Fourier transform is used to enhance the data to obtain richer details and clarity, and then the enhanced image and the original image are compared through a single-step object detection network. We pre-select the ROI area, perform segmentation and extraction based on the pre-selected ROI area, and then use histogram equalization and binarization operations to extract the outline and make a mask, use the mask to extract the ROI area of the depth map. The depth map ROI is converted into a point cloud format to obtain an approximation of reality. The shape point cloud of the value is then extracted by point cloud fitting and registration methods to obtain high-precision and stable single small object positioning and pose. The sub-diagram of the instantiated medical markup graphics processing steps is shown in Figure 2.

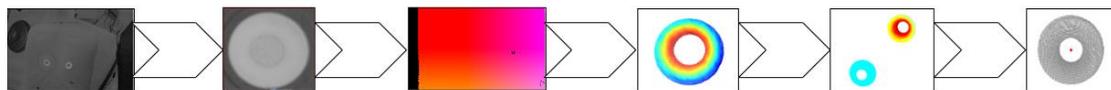

*Figure 2: Example showing image processing and feature extraction sequence. The sub-images are processed in each stage in order from left to right: the original image first undergoes Fourier transform enhancement, and is combined with the original image to become a mixed dataset adding model. Then, through the 2D object detection network segment, the mask contour is extracted and entered into the depth map to obtain the ROI point cloud. Next, fitting and robust registration are applied to extract poses and center points. Based on these analyses, specific features are identified and located in the image. Finally, the location of the feature is marked prominently.*

**Related work:**

There are generally two definition methods for the definition of 3D small objects (1) Based on relative scale: ①.The ratio of the width and height of the object bounding box to the width and height of the image is less than a certain value, and a more common ratio value is 0.1; ②.The square root of the ratio of the object bounding box area to the image area is less than a certain value, and a more common value is 0.03; ③.Small objects are defined based on the ratio between the actual covered pixels of the object and the total pixels of the image. (2) Based on absolute scale: The most common definition currently comes from the MS COCO dataset, a common dataset in the field of object detection, which defines small objects as objects with a resolution less than 32×32 pixels. The problems existing in small object detection are the few available features, high positioning accuracy requirements, the small proportion of small objects in the existing dataset, sample imbalance, small object aggregation problems, and network structure reasons.

General research ideas for small object detection: (1) Data enhancement: the simplest and most effective method to improve small object detection performance[15]. Enhancement methods include: copy enhancement, adaptive sampling, scale matching, scaling and splicing, and self-learning data Enhance. Data enhancement solves the problem of small object information, such as lack of appearance features and texture to a certain extent, improves the generalization ability of the network, and achieves better results in the final detection performance, but at the same time it brings about computational cost. Moreover, it is often necessary to optimize the object characteristics in practical applications, but improperly designed data enhancement strategies may introduce new noise. (2) Multi-scale learning: Small objects require both deep semantic information and shallow representation information, and multi-scale Learning to combine the two is an effective strategy to improve small object detection performance. Four methods currently used: ①.Image pyramid; ②.Use shallower feature maps to detect smaller objects, and use deeper feature maps to detect larger objects; ③.Multi-scale fusion network; ④.Feature Pyramid FPN (Feature Pyramid Network) multi-scale feature fusion takes into account both shallow representation information and deep semantic information, which is conducive to feature extraction of small objects and can effectively improve small object detection performance. However, existing multi-scale learning methods not only improve detection performance, but also increase the amount of additional calculations, and it is difficult to avoid the influence of interference noise during the feature fusion process. These problems make it difficult to obtain small object detection performance based on multi-scale learning. further improvement. (3) Context learning:The method based on context learning makes full use of the object-related information in the image and can effectively improve the performance of small object detection. However, existing methods do not take into account the possible lack of contextual information in the scene, and do not specifically use easy-to-detect results in the scene to assist in the detection of small objects. Future research directions generally start from the following two perspectives: ①.Construct a context memory model based on category semantic pooling to alleviate the problem of lack of context information in the current image by using the context of historical memory; ②.Small object detection based on graph reasoning, through the combination of graph model and object detection model can specifically improve the detection performance of small objects. (4) Generative adversarial learning: By mapping the features of low-resolution small objects into features equivalent to those of high-resolution objects, it can achieve the same detection performance as larger objects. Currently, it still faces two unavoidable problems: ①.Generative adversarial networks are difficult to train, and it is difficult to achieve a good balance between the generator and the discriminator; ②.The diversity of samples generated by the generator during the training process is limited, and the performance improvement after training to a certain extent is limited. (5) Anchor-free mechanism: An idea to get rid of the anchor frame mechanism is to convert the object detection task into the estimation of key points, that is, an object detection method based on key points. object detection methods based on key points mainly include two categories: corner-based detection and center-based detection. ①.Corner-based detectors predict object bounding boxes by grouping corner points learned from convolution feature maps. ②.The object detection framework based on center prediction is called Center-Net. ③.Representative point (RepPoints) detection method. This method can automatically learn the spatial information and local semantic features of the object, which improves the accuracy of small object detection to a certain extent. ④.The fully convolution one-stage object detector FCOS (Fully convolution one-stage) avoids the problem of too many hyper-parameters and difficulty in training in methods based on the anchor box mechanism. The mainstream research directions in the future include three aspects: feature fusion, context learning, and super-resolution reconstruction.

## 2. Method and material

The medical mark to be recognized is a white 24mm diameter hollow ring with a central hollow diameter of 10mm and a thickness of 3mm. The ceramic material can be imaged well under CT and has excellent color-based optical recognition and tomography-based imaging recognition performance. It can fit well on the human body surface and accurately position the function in the body surface environment. The recognition device used is a speckle method structured light camera, which can generate 2D, depth map and point cloud information and is an RGB-D information camera. The specific identification and positioning method proposed in this paper is to first obtain the 2D image and then obtain the depth map of the point cloud, so that a clear 2D image can be obtained. If the depth image of the 2D image and the point cloud information is obtained at the same time, the 2D image acquisition will obtain a noise image full of snowflake spots. This snowflake noise is actually the projection of a random speckle disparity map. Therefore, the difference between one simultaneous acquisition and two separate acquisitions lies in the speed and processing difficulty. The disparity image may be noisy, but it is faster to collect it once. The separate acquisition image is clear and easy to process, but it takes a little longer. According to the traditional algorithm, it is recommended to collect it twice. If high speed requirements are required for image detection, one-time acquisition can be used. In addition, the speed can be adjusted in the mode of image collection clarity from another angle, but the image quality will be reduced.

### 2.1 Comparative Analysis and Existing Issues with Traditional Methodologies

The steps are as follows: First, In the image pre-processing stage, this method proposed in this paper compares pre-processing filters, high-pass filtering, histogram equalization, median filtering, mean filtering, and Gaussian filtering. Each has its own advantages and disadvantages in the comparison, but they all lead to the same results. The local effect is good and the other situation is unstable, so it is relatively recommended that histogram equalization and Gaussian filtering be used as noise reduction methods. The use of multi-classifier extraction is the core of this method. The second is to use open operation; the third is to use threshold segmentation; the fourth is to use connected areas; the fifth is to use area screening; and the sixth is to use circular similarity analysis. The seventh is obtain the 3Dxyz value center based on the fitted circle center mapping depth map. Circle extraction is based on the least squares calculation: the formula is as follows. For specific calculations, refer to the extraction formula below.

$$\int = \sum \left( (x_i)^2 + (x_c)^2 \right) + (y_i)^2 + (y_c)^2 ) - R^2 )^2$$

But this is only the algorithm stage. In practice, external assistance is very

important. The steps are:

Step 1: Optical stability light environment observation. First of all, there is a light environment parameter table in the medical environment. Secondly, various on-site light environment sampling.

Step 2: Contrast enhancement of the observed object. It is a good way to use a surrounding black edge ring with strong black and white contrast to highlight the marker, which can ensure stable and clear extraction. The material selection of this contrast object is very important. The mirror metal material is not recommended for selection, because it seems that the light perception contrast is very strong, but in a certain some angles have high contrast, but some angles have very low contrast. To achieve good optical contrast enhancement materials, it is recognized that the best light-absorbing material is carbon nanotubes. However, this material is not only expensive and has a long cycle, but also has the characteristics of being difficult to obtain. Therefore, the simple and easy solution used in the experiment is to use matte material, which is often a rough and porous surface. In practice, the optical performance of medical scene markers is almost the same.

Step 3: Limit the scope of observation and the angle of observation. The observation angle is limited. It is normal for the observation angle to be within 30 degrees. The recognition deviation is large at an excessively large angle, so it is better to limit it to a certain angle. Otherwise, the set circle fitting deviation including ellipse fitting will have poor fitting. For example, it will be very stable if the observation is limited to a relative angle of 30 degrees and a height of 400mm. The observation effect is shown in Figure 3.

It can be seen that this method can only observe at a small angle and has poor stability, and is also sensitive to ambient light. The advantage is that it is simple to deploy, lightweight and quick to use, but it is difficult to avoid the problems of depth map mapping loss and optical blur.

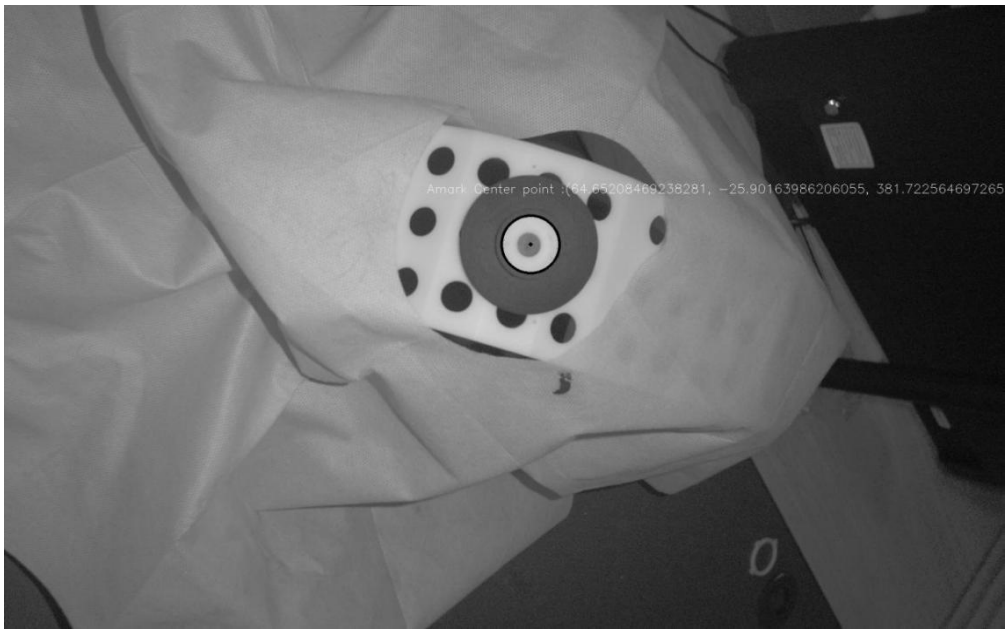

*Figure 3: Recognition and positioning effect diagram. It shows the stable recognition status that*

*limited to a relative angle of 30 degrees and a height of 400mm.*

The optical blur value will cause the coordinate positioning accuracy to fluctuate below the stable sub-millimeter level. Depending on the product, it will be between 0.2-0.7mm. The ideal observation posture and observation distance are one solution, but it can only avoid part of it and limit the scope of flexible application in practice. For problems based on optical blur coefficients such as light interference, the optical blur changes with distance, as shown in Figure 4 below.

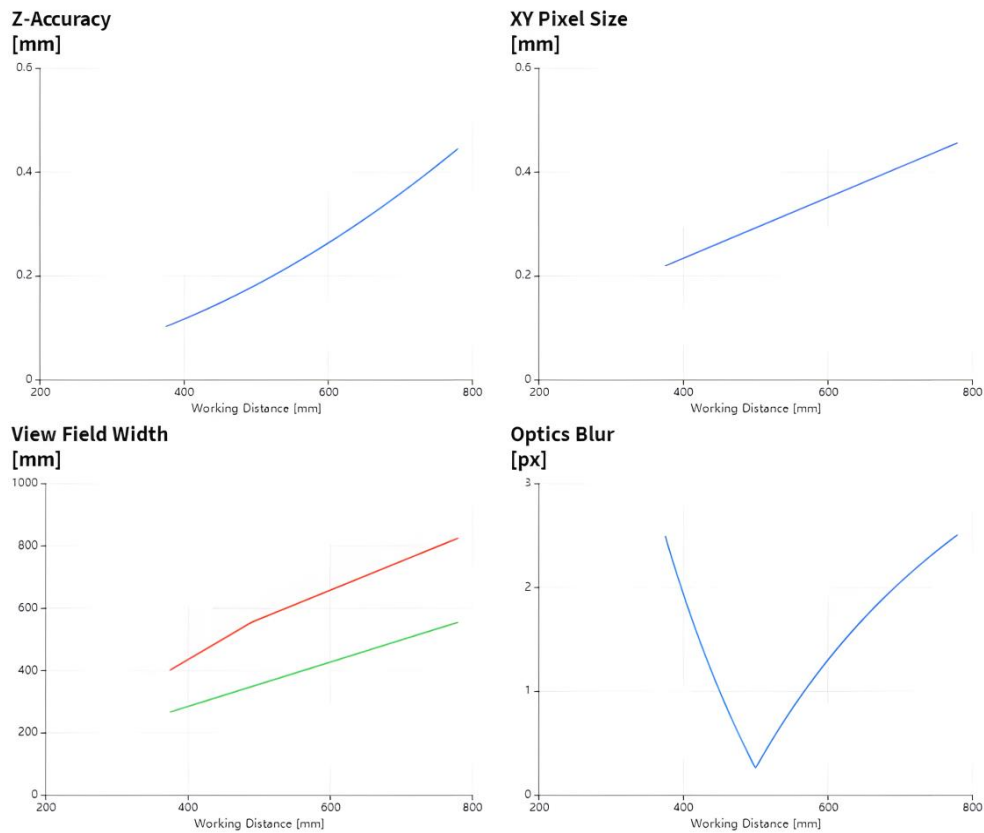

*Figure 4: The trends of four different measurement parameters as a function of working distance. Each parameter is plotted in its own subplot.*

*Z-Accuracy [mm]: This image description shows the trend of Z-axis (depth) accuracy gradually getting worse with increasing working distance. As the working distance increases, the depth accuracy error of the measurement also increases.*

*XY Pixel Size [mm]: This graph depicts the tendency of pixel size in the XY plane to increase with increasing working distance. The farther the working distance, the larger the actual area covered by each pixel, resulting in reduced resolution.*

*View Field Width [mm]: This graph shows the linear increase in field width with increasing working distance. This means that as the camera moves further away from the subject, its field of view covers a wider area.*

*Optics Blur [px]: This graph shows the optical blur (in pixels) trend, which reaches a minimum value within a certain working distance range, and then increases again as the working distance continues to increase. This indicates that there is an optimal working distance at which the optical system provides the sharpest image.*

If direct RANSAC and DBSACN point cloud fitting are omitted, the instability will increase due to the sparse point cloud density at different distances. Algorithms such as RANSAC themselves are called uncertainty algorithms, resulting in unstable extraction or failure. Common methods of point cloud fitting often have uncertainties.

The RANSAC algorithm itself is a non-deterministic algorithm. The clustering algorithm will also become less stable as the point cloud sparse values are used. Other effective methods such as arbitrary surface cutting circles are used. Ring extraction and non-directional random cutting will lose the direction accuracy, lose the original intention of improving accuracy, and become meaningless.

## 2.2 Fourier-enhanced multi-modal point cloud refinement method

For structured light cameras, point clouds are characterized by collecting multiple disparity maps. The information is richer than a single shot of 2D images. Compared with optics, it is more accurate in producing camera optical blur values at different distances and can also avoid depth mapping. Mapping loss occurs because this method is not used, so the point cloud extraction effect in this scene is theoretically more ideal. However, on-site extraction of small object point clouds of 24mm rings is a challenge.

Based on this, a multi-modal method of using fast Fourier transform to enhance image details is proposed. Indeed, enhancing the resolution and meticulously extracting the predefined image contour, followed by the retrieval of a 24-60mm point cloud, facilitates the refinement of point cloud fitting and registration, thereby elevating precision. This approach circumvents the pitfalls of optical blur and depth map mapping loss. Moreover, it exhibits superior resilience to optical disturbances, ensuring the stability and accuracy in feature enumeration and classification—a quintessential attribute for applications demanding medical-grade precision and stability.

First, we use fast Fourier transform to enhance the image details, and then use a single-step object detector (such as YOLO, SSD, etc.) combined with the depth map mapping ROI method. In this paper, we adopt YOLOv5 as the baseline model to extract the region of interest (ROI) through a single-step object detection network and the identified ROI is subsequently segmented from the image. Second, a BLOB operation is conducted on the segmented area to delineate its contours, which are then rendered onto an empty array corresponding to the original image dimensions, creating a mask image. This mask facilitates the cropping of the depth map, yielding the target area's ROI depth map. The depth map is then transformed into a point cloud, producing the target point cloud. Fitting extraction based on the point cloud's shape and registration culminates in the retrieval of precise, stable, and minimally interfered position information for small targets. Third, We determine the center point of the target based on the size of the rectangular frame, and use the coordinates of the center point to find the corresponding mapping point in the depth map to obtain the coordinate value of the target point. It is worth noting that the single-step detection network usually processes three-channel images. For single-channel images, they need to be converted into three-channel images. When the target cannot be recognized, adding unrecognized samples for retraining can solve the problem. In practice, sparse sample situations are extremely common. By augmenting the dataset and fine-tuning the model, the model performance can ultimately be improved to a higher level. The specific processing flow is shown in Figure 5.

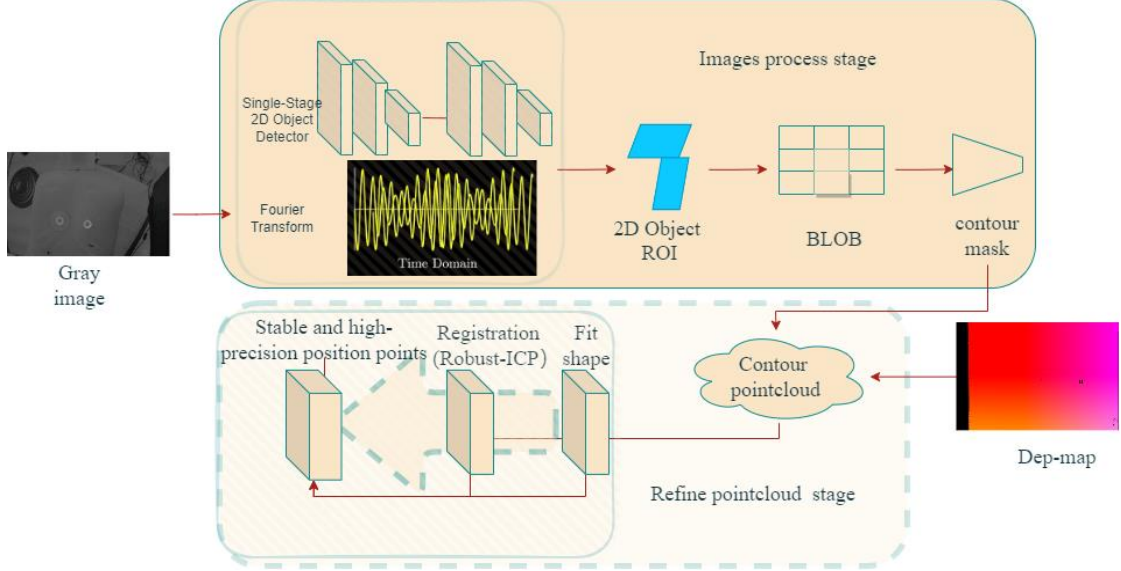

*Figure 5: The grayscale image and the depth map are collected twice. They have the same size and mapping relationship. Gray is the grayscale image, and red is the depth map. The grayscale image is enhanced by Fourier transform, and the data is not enhanced. The sets form a hybrid enhanced dataset and are trained together in the model. The trained model is used to infer the grayscale image and generate a 2D target area. Then BLOB analysis is performed to obtain the contour mask to obtain the cropping depth map, and the contour point cloud is generated based on the contour depth map. Finally, it can obtain stable and accurate positioning points through a robust registration method with geometric shape fitting with Tukey loss.*

### 2.2.1 Data acquisition and pre-processing

A high-quality dataset is the foundation and premise for a superior model, crucial for enhancing optical performance and point cloud effects. In private domain scenarios, we collected and created a diverse range of real-shot data under laboratory and hospital CT room environments, featuring various angles and simulated situations. Adhering to medical building standards, the lighting conditions in our experimental setups ranged from 130 to 750LX, incorporating common environmental interference tools during data collection. We captured 2D and 3D images using structured light cameras to accommodate multi-modal experiments, thus employing a fusion labeling approach for both 2D and 3D data. Specifically, 350 2D images were annotated using the LabelImg tool, while 330 multi-modal fused 3D point clouds and 370 point cloud and image fusion samples were annotated with the Xtreme1 tool. To expand the dataset in later stages, we utilized model-based line data augmentation methods, potentially increasing the dataset size by 2 to 5 times.

## 2.2.2 Fast Fourier transform enhancement

Performance on images is often affected by the aspect ratio being too small and the image pixels not being obvious. It is a common practice to use image enhancement and linear difference. image enhancement methods include common contrast sharpening, contrast enhancement, brightness enhancement, etc. However, image enhancement will also bring about the problem of larger low-order values and poor stability. Fast Fourier transform is a good method, but it has not been applied to 3D small objects before. Fast Fourier transform (FFT) is a An algorithm in digital signal processing changes in the frequency domain: Fourier transform splits the signal in the spatial domain into signals in different frequency bands, and converts the pixel image into a spectrogram.

The idea of frequency domain detection is to detect from the spatial domain to Frequency domain, perform appropriate filtering in the frequency domain, select the frequency band you want, and then return to the spatial domain. There are two key factors, one is to generate a suitable filter, and the other is the conversion between the spatial domain and the frequency domain. This paper uses Gaussian filtering. The sigma value of Gaussian filtering is the standard deviation of the Gaussian kernel function, which determines the width and shape of the Gaussian kernel function. The larger the sigma value, the wider the width of the Gaussian kernel function, and the smoother the image; the smaller the sigma value, the narrower the width of the Gaussian kernel function, and the more prominent the details of the image. The method is to set two Gaussian filter kernels, one with a sigma value equal to 3 and the other with a sigma value equal to 15. After passing through two Gaussian filters respectively, the two images are subtracted, so that they are converted to frequency domain processing, and then after converting it back, we will find that the image with a very low threshold difference becomes a large gap and the outline is clear and obvious. The pseudo code is as follows:

Algorithm 1: Enhanced 3D Object Recognition using Fourier Transform and Gaussian Filters

Input: Directory of image files

Output: Display images and image variable

1: for each image in the directory do

2: # Convert image to grayscale

3: # Apply Fourier Transform to convert the image to frequency domain

4: fft_generic(GrayImage, ImageFFT, 'to_freq', -1, 'n', 'dc_center', 'complex')

5:

6: # Generate Gaussian filters for background weakening and defect enhancement

7: gen_gauss_filter(ImageGauss, 15, 15, 0, 'none', 'dc_center', Width, Height)

8: gen_gauss_filter(ImageGauss1, 3, 3, 0, 'none', 'dc_center', Width, Height)

9:

10: # Subtract the generated Gaussian filters to highlight defects

11: sub_image(ImageGauss1, ImageGauss, ImageSub, 1, 128)

12:
13: # Apply convolution in the frequency domain using the subtracted image
14: convol_fft(ImageFFT, ImageSub, ImageConvol)
15:
16: # Convert the convoluted image back to the spatial domain
17: fft_generic(ImageConvol, ImageFFT1, 'from_freq', 1, 'n', 'dc_center', 'real')
18:
19: # Analyze the image for object recognition (eg, Blob analysis)
20: Perform Blob Analysis on ImageFFT1
21:
22: Display the enhanced image

How to view the spectrogram description (information includes: phase, frequency, and frequency band content):

(1) Each point is not a pixel, but represents a different frequency or frequency band, and the center of the spectrogram is low frequency, which gradually increases from the center to the surroundings. Sometimes the center of the spectrum is not at the center of the window.

(2) The spectrum chart is symmetrical up, down, left and right.

(3) The brightness of each point in the spectrogram represents the amplitude of the frequency band (and the degree of grayscale transformation). If the bright spot has a certain width, it means that the frequency content at this point is relatively rich.

(4) There are bright spots or bright lines in a certain direction in the spectrogram, indicating that there are edges with drastic grayscale changes in that direction.

### 2.2.3 Training details

We use a GeForce RTX 4080 for training, and the batch size is set to 128. Usually each training is 500 epoch, and the accuracy rate is 99.8% as the inference model. Two parameters that are often adjusted are ROI pooing and Anchor box.

Setting of hyperparameters: In the loss design, $\alpha$ and $\gamma$ focal loss are set to [0.75, 0.75, 0.25] and [2., 2., 2.] respectively. The three loss weights are set to $\lambda_{loc} = 1$, $\lambda_{cls} = 1$ and $\lambda_h = 1.5$. The Adma optimizer[16] is used to train for 70 epochs with an initial learning rate of $lr = 2e^{-4}$, and the weight decay is set to $1e^{-4}$. We first adopt the warm-up strategy[17], with 300 warm-up iterations and a warm-up ratio of 1./3 . In addition, the decay ratio of lr at the 40th epoch and 60th ecpoch is 0.1. During the inference phase, the prediction score threshold is set to 0.2 and the rotated NMS[18] threshold is set to [0.02, 0.02, 0.4].

Configuration of single-step detector on data augmentation: flip lr: 0.5 (probability of left and right flipping); mosaic: 1.0 (probability of mosaic data enhancement ); mix up: 0.0 (probability of mixed images); hsv_h: 0.015 (probability of hue dithering); hsv_s: 0.7 (probability of saturation dithering) ; hsv_v: 0.4 (probability of luminance dithering); degrees: 0.7 (angle range of rotation); translate: 0.1 (panning range); scale: 0.5 (zooming range) , which can be increased by 2-3 times by configuring the actual image.

**2.2.4 The process of masking the ROI obtained by cropping the depth map:**

It is to create an empty array with the same size as the depth map, and then only have the mask map in it. Set all the mask numbers to 1, and the remaining blank parts to 0. Then the mask map is actually a multi-dimensional array, which is also a matrix. Multiplying the array (matrix) of the depth map, you get the mask map. The mask part is the depth map, the depth information part, and the rest of the non-roi area is 0. Then remove all the 0 values in this array, and you get The depth map is cropped by the mask image. This depth map is the depth map of the area of interest, and the rest has been deleted.

## 2.3 Point cloud refinement: fitting and registration

This paper adopts the refinement part of the progressive method of two superposition classifiers, and obtains strong robustness and high accuracy results through fitting and registration superposition combination. Since blob analysis is used to extract contours, it is often a fixed value, resulting in the extracted point cloud containing noise information in a few cases. Therefore, the fitting will lead to an increase in error, so a registration is added to improve the accuracy, because the registration is Extracted from the standard template, coupled with the effect of Tu-based loss, it has less impact on noise. In fact, fitting can extract the center of the circle. However, in order to prevent the error from increasing and unstable recognition, registration is added. These two are in a progressive relationship. If the fitting and registration errors are large enough, they will exit to ensure accurate positioning and small errors every time during the operation. This is due to the high stability and accuracy requirements of surgical positioning.

In the process of extracting ROI point clouds, Gaussian filtering and binarization are initially applied to obtain the ROI contours, which are then used to generate a new image serving as an image mask. This mask can also be created by drawing contours. Subsequently, this mask is utilized to crop the depth map, producing the ROI depth map. For depth maps acquired with structured light cameras, this method can be directly applied. In cases where depth maps are captured using LiDAR and cameras, multi-camera calibration is necessary to establish the mapping relationship, typically requiring mapping to images of the same size. However, employing bitmap methods for mask cropping is impractical because such methods usually operate on UNIT8, and converting depth maps to UNIT8 format would result in loss of depth information. Instead, multi-loop techniques or array multiplication can be used for mask mapping, facilitating the extraction of the ROI from the depth map. Afterward, the depth map can be converted into a point cloud for subsequent refinement operations on the ROI point cloud. By removing noise and environmental information, the target point cloud is simplified into a simple annular shape, making RANSAC fitting easier. Direct extraction with RANSAC at the outset may fail due to small target size or environmental noise. In addition, Point cloud information is often richer than image information and is more stable and accurate for collecting precise locations.

### 2.3.1 Point cloud volume fitting method

Extracting refined point clouds is relatively simple, accurate, and stable. Since the mark is a donut shape, it is connected to the environment to form a cone. Therefore, the shape of the point cloud can be relatively ideal using the cone fitting. The specific method Refer to the cone fitting formula of the following fitting:

$$\frac{\sqrt{(x - x_o)^2 + (y - y^o)^2}}{z - z_o} = \tan(\theta)$$

Explanation: (1) Geometric properties of the cone: For a cone with (x0, y0, z0) as the vertex and the axis aligned with the z-axis, any point (x, y, z) located on its surface satisfies the above relationship, here, the numerator is the horizontal distance from the point (x, y, z) on the xy plane to the vertex of the cone, and the denominator is the vertical distance from the point to the vertex. This ratio is equal to the tangent of the cone's opening angle. (2) Calculate the theoretical z coordinate: According to the above formula, we can rearrange the formula to calculate the theoretical z coordinate. The pseudo code is as follows:

# Algorithm 2 :      Calculate the theoretical z using the cone equationtan
              Tan_theta = np.tan(theta)
              Theoretical_z = z0 + np.sqrt((x- ×0)**2 + (y - y0)**2)/tan_theta

Where, theoretical_z is the predicted z coordinate value based on the cone model.
(3) Calculate the residual. The residual is the difference between the z coordinate of each point in the actual point cloud data and the theoretical predicted value. During the optimization process, the goal is to minimize the sum of squares of these residuals for all points.

### 2.3.2 Tukey loss registration method

ICP (Iterative Closest Point) has many registration types. For medical labeling scenarios, a robust registration method with high stability is required. This paper uses the Tukey loss robust registration method, which has good robust effects. Meanwhile, it can output very stable results for the point cloud registration method. The formula is as follows:
The loss $\rho(r)$ for a given residual r is:

$$\rho(r) = \begin{cases} \dfrac{k^2\left[1-\left(1-((1-(\frac{e}{k})^2)^3\right)\right]}{2}, & |r| \le k. \\ \\ \dfrac{k^2}{2}, & otherwise \end{cases}$$

The weight *w(r) for given residual* r *is given by：*

$$w(r) = \begin{cases} \left(1-\left(\dfrac{r}{k}\right)^2\right)^2, & |r| \le k. \\ \\ 0, & otherwise \end{cases}$$

this formula is the Tukey loss function and its corresponding weight function, which are used to calculate the loss value and weight of the residual.

When the residual r is less than or equal to the threshold k, the shape of the loss function $\rho(r)$ is similar to the square loss, but when the absolute value of the residual increases to close to k, the growth rate slows down and remains unchanged after exceeding k. This design reduces the weight of large residuals in the total loss, thereby improving robustness to outliers.

The weight function *w(r)* is the derivative of the loss function to the residual r, which is used to adjust the influence of each data point during the iterative optimization process. When the residual is less than or equal to k, the weight decreases as the residual increases; when the residual is greater than k, the weight is zero, that is, these possible outliers are ignored. This helps the optimization process focus more on data points that fit the model.

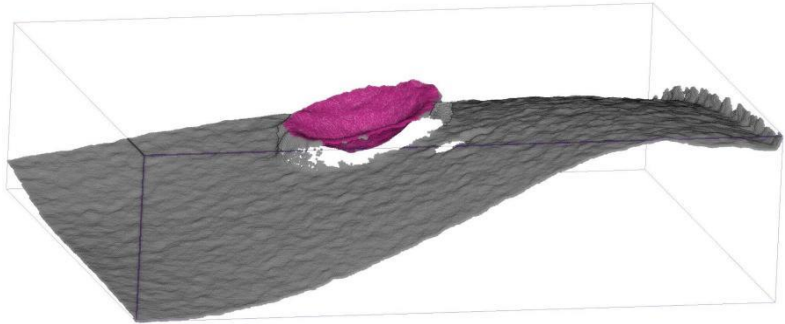

*Figure 6: ROI point cloud obtained from the whole scene, the purple point cloud segmentation part, and the gray part is the background part.*

The specific method is to use the estimated normal vector method to calculate the normal for the two point clouds, which is necessary for the ICP alignment of the point to the plane. The initial alignment uses the identity matrix np.eye(4) as the initial transformation. This means that the source and object point clouds are initially aligned without any rotational or translational alignment. Point-to-plane ICP alignment, use the point-to-plane ICP method to initially align the source point cloud and the object point cloud, apply the resulting transformation to a copy of the source point cloud and display the result, and then try to use the Tukey loss function Variant ICP method. The Tukey loss function is a robust loss function that is less sensitive to outliers and works well for processing data containing noise or outliers. The result after segmentation is shown in Figure 6.

## 3. Result

Accuracy is a multiple factor, including algorithms, mechanical assembly, ground vibration, CT bed displacement, etc. It can be analyzed by type, but it is difficult to judge alone because it still needs to be experienced in the actual environment, especially in the CT room where there are constant under the condition of heavy equipment operation, the before and after algorithms can be compared with the numerical values under actual measured vibration.

Table 1: Comparative acquisition accuracy is as follows, previously using depth map mapping method

| Traditional mapping algorithm places A-Mark recognition results | | | |
|---|---|---|---|
| serial number | Recognition results | | Straight line distance difference (±) |
| | XYZ | | |
| 1 | 60.36693 | -4.17822 | 363.7445 | 1.332731176 |
| 2 | 61.35583 | -2.3430135 | 364.15797 | 0.821445875 |
| 3 | 60.34972 | -4.1770315 | 363.6408 | 1.369640695 |
| 4 | 60.4186 | -3.2620718 | 364.05582 | 0.515140576 |
| 5 | 61.39088 | -2.344352 | 364.36777 | 0.880754035 |
| 6 | Recognition failed | | |
| 7 | 60.36693 | -4.178227 | 363.7445 | 1.33273741 |
| 8 | 61.408417 | -2.3450217 | 364.47183 | 0.927111199 |
| 9 | 61.39088 | -2.344352 | 364.36777 | 0.880754035 |
| 10 | 60.34972 | -4.1770315 | 363.6408 | 1.369640695 |
| 11 | Recognition failed | | |
| 12 | 61.373352 | -2.3436825 | 364.2637 | 0.845102971 |
| 13 | 60.34972 | -4.1770315 | 363.6408 | 1.369640695 |
| 14 | 60.487637 | -2.3450217 | 364.47183 | 0.828496361 |
| 15 | Recognition failed | | |
| 16 | 61.408416 | -2.3450217 | 364.47183 | 0.927110601 |
| 17 | 60.384144 | -3.2602117 | 363.8482 | 0.599204743 |
| mean | 60.85401488 | -2.991194841 | 364.1049053 | **0.973682037** |

| | | | |
|---|---|---|---|
| Standard deviation σ | 0.501098721 | 0.817832392 | 0.314885796 |

Table 2: Actual performance test table using this algorithm

| Multi-modal point cloud refinement placement A-Mark recognition results | | | |
|---|---|---|---|
| serial number | Recognition results | | Straight line distance difference (±) |
| | XYZ | | |
| 1 | 57.93445 | 1.32185 | 360.04922 | 0.10323271 |
| 2 | 57.95081 | 1.32222 | 360.15085 | 0.136184428 |
| 3 | 57.93445 | 1.32185 | 360.04922 | 0.10323271 |
| 4 | 57.93445 | 1.32185 | 360.04922 | 0.10323271 |
| 5 | 57.93445 | 1.32185 | 360.04922 | 0.10323271 |
| 6 | 57.53772 | 1.14727 | 360.06187 | 0.716752595 |
| 7 | 57.93445 | 1.32185 | 360.04922 | 0.10323271 |
| 8 | 57.91811 | 1.32147 | 359.94763 | 0.154767426 |
| 9 | 57.87316 | 1.23348 | 360.35431 | 0.285762242 |
| 10 | 57.73445 | 1.32185 | 360.04922 | 0.110382359 |
| 11 | 57.93445 | 1.32185 | 360.04922 | 0.10323271 |
| mean | 57.83826818 | 1.297944545 | 360.0781091 | **0.133580156** |
| Standard deviation σ | 0.121477809 | 0.053937284 | 0.097565242 | |

Through the above operation algorithm, the repetition accuracy is identified from the standard deviation value of the accuracy of **0.973682037** mm to the repetition accuracy error value of **0.133580156**. Recognition was originally unrecognizable due to misoperation, but it will no longer occur after multi-modal point cloud refinement is used.

## 4. Discussion

The Fourier-enhanced multi-modal point cloud refinement method proposed in this paper provides a novel solution for 3D small object detection at the medical labeling level. Although existing 3D object detection methods have achieved significant progress on large objects, but the detection methods of small objects, especially small medical markers for use in surgical navigation, are still under development. This paper effectively improves the detection accuracy and stability of small objects in this application case by combining a single-step object detection network and point cloud processing technology.

This method has a positive impact on medical mark recognition. Previously, the recognition method relied on optical reflective balls, which were difficult to fit on the human body surface. There was a level of positioning of surgical instruments. For body surface points, this method predominantly used three reflective balls, posing numerous challenges in their placement. Inferring from a certain distance from the epidermis further complicated matters. However, the use of white ring markers can accurately reflect skin points, particularly beneficial for surgeries requiring precise fitting, such as those involving the chest and abdomen. This method has undergone practical testing and is point-based.

Based on point cloud 3D general neural networks are mostly for large objects, making the application of this small object method still relatively uncommon in practice. Meanwhile, this method will increase the amount of calculation and the calculation time, but the overall time is influenced by various factors due to the robustness of the algorithm. Therefore, it is feasible to process lower quality images and sparse point clouds, which can offset the time required for camera acquisition.

Cameras are generally divided into 5 modes from high to low based on parameter adjustments (levels of quality and speed): static quality, static balance, static fast, dynamic balance, and dynamic fast. The difference is the clarity of captured images and the frequency of projected images. Higher modes provide faster imaging but with relatively more blur, resulting in larger accuracy errors. Conversely, slower modes yield better quality and higher accuracy. Previously, the acquisition time required in static balance and static quality modes was 1.7 and 2.1 seconds. However, using the dynamic fast mode can complete acquisition in just 0.5 seconds. Hence, by enhancing the graphics card configuration and adjusting camera speed settings, this time can be significantly reduced.

The main advantage of this method is its high-precision positioning capability for small objects. Traditional 3D object detection methods often suffer from performance degradation when dealing with small-sized objects due to insufficient features and high positioning accuracy requirements. By utilizing 2D images to extract ROI areas and combining them with point cloud technology, this method can handle small-sized objects more accurately, which has positive significance for surgical navigation.

However, this paper also has some limitations. Firstly, the accuracy of the method is highly dependent on the quality of the 2D image and the accuracy of the mask. Any errors during the 2D image processing stage may lead to a decrease in the accuracy of the final point cloud. Secondly, the fitting and registration processes in the point cloud processing step require precise tuning and optimization to ensure optimal performance. Despite these challenges, the method proposed in this paper showed excellent performance in experiments. Especially in repeatability tests, this method can significantly reduce positioning errors, which is extremely important for practical surgical applications.

## 5. Conclusion

This paper successfully developed a multi-modal point cloud refinement method for 3D small object detection, particularly focusing on the precise positioning of medical markers. For medical surgical navigation, body surface positioning is found to be more suitable for accurately placing needles on lesion areas of the body surface. This method combines fast Fourier transform, single-step object detection network and point cloud technology to effectively improve the detection accuracy and stability of small objects in complex medical environments. Detailed experiments and analysis demonstrate the practicality and effectiveness of this method in medical surgical navigation.

This method can position any needle insertion point based on the body surface. In the future, it holds promise for applications in body surface positioning for puncture ablation across diversified surgical scenarios. As a novel technique in the field of surgical navigation, it has positive significance for puncture ablation surgical navigation of tumors.

In summary, this paper provides an effective method for 3D small object detection in surgical navigation, and contributes positively towards enhancing the safety and accuracy of thoracoabdominal surgical navigation.

## Acknowledgement

This work was supported by the Hebei Provincial Social Science Foundation Project (No.HB20TQ003). We thank Beijing Yingrui Pioneer Medical Technology Company for experimental data support and Professor Zhao Zhanfang of Hebei GEO University for correcting the writing of this article.

## Reference

[1] CR Qi, H. Su, K. Mo, and LJ Guibas. Pointnet: Deep learning on point sets for 3d classification and segmentation. In CVPR, pages 652–660, 2017.

[2] CR Qi, L. Yi, H. Su, and LJ Guibas. Pointnet++: Deep hierarchical feature learning on point sets in a metric space. In NeurIPS, pages 5099–5108, 2017.

[3] B. Graham, M. Engelcke, and L. Van Der Maaten. 3d semantic segmentation with submanifold sparse convolutional networks. In CVPR, pages 9224–9232, 2018.

[4] C. Choy, J. Gwak, and S. Savarese. 4d spatio-temporal convnets: Minkowski convolutional neural networks. In CVPR, pages 3075–3084, 2019.

[5] S. Shi, C. Guo, L. Jiang, Z. Wang, J. Shi, X. Wang, and H. Li. Pv-rcnn: Point-voxel feature set abstraction for 3d object detection. In CVPR, pages 10529–10538, 2020.

[6] W. Zheng, W. Tang, L. Jiang, and C.-W. Fu. Se-ssd: Self-ensembling single-stage object detector from point cloud. In CVPR, pages 14494–14503, 2021.

[7] H. Wang, L. Ding, S. Dong, S. Shi, A. Li, J. Li, Z. Li, and L. Wang. Cagroup3d: Class-aware grouping for 3d object detection on point clouds. arXiv preprint arXiv:2210.04264, 2022.

[8] D. Rukhovich, A. Vorontsova, and A. Konushin. Tr3d: Towards real-time indoor 3d object detection. arXiv preprint arXiv:2302.02858, 2023.

[9] A. Geiger, P. Lenz, and R. Urtasun. Are we ready for autonomous driving the kitti vision benchmark suite. In CVPR, pages 3354–3361, 2012.

[10] A. Dai, AX Chang, M. Savva, M. Halber, T. Funkhouser, and M. Nießner. Scannet: Richly annotated 3d reconstructions of indoor scenes. In CVPR, pages 5828—-5839, 2017.

[11] A. Chang, A. Dai, T. Funkhouser, M. Halber, M. Niessner, M. Savva, S. Song, A. Zeng, and Y. Zhang. Matterport3d: Learning from rgb-d data in indoor environments . 3DV, 2017.

[12] I. Armeni, O. Sener, AR Zamir, H. Jiang, I. Brilakis, M. Fischer, and S. Savarese. 3d semantic of large-scale indoor spaces. In ICCV, pages 1534–1543, 2016.

[13] S. Song, SP Lichtenberg, and J. Xiao. Sun rgb-d: A rgb-d scene understanding benchmark suite. In CVPR, pages 567–576, 2015.

[14] X. Xu, Y. Wang, Y. Zheng, Y. Rao, J. Zhou, and J. Lu. Back to reality: Weakly-supervised 3d object detection with shape-guided label enhancement. In CVPR, pages 8438– 8447, 2022.

[15] 1 Rabbi, J.; Ray, N.; Schubert, M.; Chowdhury, S.; Chao, D. Small-Object Detection in Remote Sensing Images with End-to-End Edge-Enhanced GAN and Object Detector Network. Remote Sens. 2020, 12, 1432. https://doi.org/10.3390/rs12091432

[16] Diederik P Kingma and Jimmy Ba. Adam: A method for stochastic optimization. arXiv preprint arXiv:1412.6980, 2014. 6

[17] Kaiming He, Xiangyu Zhang, Shaoqing Ren, and Jian Sun. Deep residual learning for image recognition. In Proceedings of the IEEE conference on computer vision and pattern recognition, pages 770–778, 2016. 3, 6

[18] Alexander Neubeck and Luc Van Gool. Efficient nonmaximum suppression. In 18th International Conference on Pattern Recognition (ICPR'06), volume 3, pages 850–855. IEEE, 2006. 6